\documentclass[sigconf]{acmart}

\AtBeginDocument{%
  }

\setcopyright{acmcopyright}
\copyrightyear{2018}
\acmYear{2018}
\acmDOI{XXXXXXX.XXXXXXX}

\usepackage[ruled]{algorithm2e}
\usepackage{multirow}

\begin{document}

\title{SLIC: Secure Learned Image Codec through Compressed Domain Watermarking to Defend Image Manipulation}


\author{Chen-Hsiu Huang}
\email{chenhsiu48@cmlab.csie.ntu.edu.tw}
\orcid{0000-0003-4352-0375}
\affiliation{%
  \institution{National Taiwan University}
  \city{Taipei}
  \country{Taiwan}
}

\author{Ja-Ling Wu}
\orcid{0000-0002-3631-1551}
\email{wjl@cmlab.csie.ntu.edu.tw}
\affiliation{%
  \institution{National Taiwan University}
  \city{Taipei}
  \country{Taiwan}
}


\begin{abstract}
The digital image manipulation and advancements in Generative AI, such as Deepfake, has raised significant concerns regarding the authenticity of images shared on social media. Traditional image forensic techniques, while helpful, are often passive and insufficient against sophisticated tampering methods. This paper introduces the Secure Learned Image Codec (SLIC), a novel active approach to ensuring image authenticity through watermark embedding in the compressed domain. SLIC leverages neural network-based compression to embed watermarks as adversarial perturbations in the latent space, creating images that degrade in quality upon re-compression if tampered with. This degradation acts as a defense mechanism against unauthorized modifications. Our method involves fine-tuning a neural encoder/decoder to balance watermark invisibility with robustness, ensuring minimal quality loss for non-watermarked images. Experimental results demonstrate SLIC's effectiveness in generating visible artifacts in tampered images, thereby preventing their redistribution. This work represents a significant step toward developing secure image codecs that can be widely adopted to safeguard digital image integrity.
\end{abstract}




\maketitle

\section{Introduction}

Digital image manipulation has long posed a security threat to images distributed on social media. Nowadays, people no longer trust sensitive images spreading on social networks and often need to re-confirm the image source. Detecting forged images from simple editing operations such as splicing and copy-moving are challenging tasks. The advent of Deepfake \cite{tolosana2020deepfakes} and text-to-image diffusion models \cite{zhang2023text} in the Generative AI era has further exacerbated the credibility issue of social media images, as tampering and counterfeiting images are no longer restricted to experts. These altered images are re-encoded in either lossless or lossy formats. As the most popular lossy image codec, JPEG exhibits specific characteristics in its DCT coefficients, and there are methods \cite{mahdian2009detecting,park2018double} to detect double JPEG compression to identify manipulated images. For lossless encoded images, pixel-based characteristics such as noise \cite{mahdian2009using}, patterns \cite{bayram2009efficient}, and camera properties \cite{ghosh2019spliceradar,popescu2005exposing} are analyzed to detect local inconsistencies and identify image forgery.

\begin{figure}[!t]
\centering
\includegraphics[width=1\columnwidth]{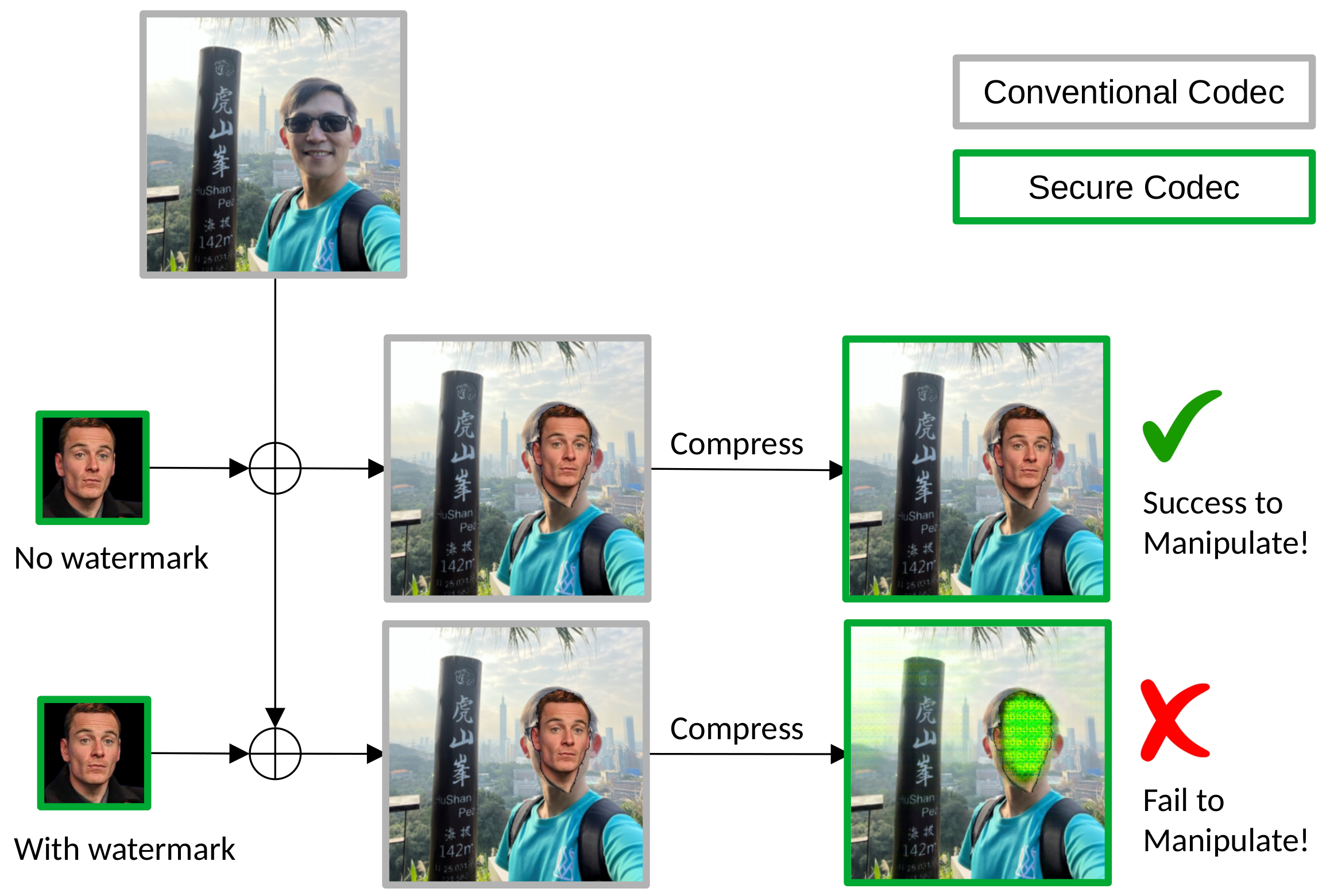}  \\
\caption{The proposed Secure Learned Image Codec (SLIC) validates content authenticity by severely degrading the image content if the source image contains any protected content. Embedding a watermark in the latent representation can be thought of as perturbations to generate adversarial examples for a neural compressor.}
\label{fig:sample}
\end{figure}

Image forensics techniques \cite{piva2013overview,zanardelli2023image} are considered passive approaches to proving content authenticity because they detect tampering after it has occurred. Active approaches such as trustworthy cameras \cite{friedman1993trustworthy,blythe2004secure} or fragile digital watermarking \cite{kundur1999digital,lu2000structural} attempt to authenticate the image after its creation and detect any later modification by checking the embedded signature or watermark. We propose a novel active approach, the Secure Learned Image Codec (SLIC), to validate that contents in the SLIC format do not contain any watermarked source. Figure \ref{fig:sample} shows our proposed application scenario, which forces a malicious party to obtain severely damaged visual content after re-compressing a spliced image in the SLIC format. The owner of sensitive images can publish them in the SLIC format with a watermark to prevent unauthorized editing. Since the content of a SLIC image is either original or authorized by the content owner without a watermark, a secure codec-encoded image has more credibility than other images. If the public tends only to trust SLIC images, then malicious parties will have to produce fake images in SLIC format to gain trust, which is considered a non-trivial task. Thus, we can prevent counterfeiting behavior from the beginning.

Designing a secure codec based on conventional codecs is challenging because most of them use linear and unitary transforms such as the Discrete Cosine Transform (DCT). The non-linear transform used in a learned image codec has the potential for ``destroy after re-compression'' because adversarial examples are known to neural networks. Adversarial attacks add invisible perturbations to the input image to cause file size expansion \cite{liu2023manipulation} or severe quality degradation \cite{chen2023towards}. We believe there are two approaches to developing a secure learned codec:

\begin{enumerate}
\item Train a neural codec whose decoder will generate out-of-distribution perturbations not seen in the training dataset.
\item Leverage a compressed domain data hiding \cite{huang2024joint} technique so that the message added in the latent vector will cause out-of-distribution perturbations in the reconstructed image.
\end{enumerate}

To the best of our knowledge, we are the first to propose the concept of a secure image codec via learned image compression using these two possible approaches. The main contributions of this paper are summarized as follows:

\begin{itemize}
\item We introduce SLIC, a novel active approach to prove content authenticity using a secure learned image codec. SLIC can effectively defend against unauthorized image manipulation through watermark embedding, which generates adversarial examples that degrade visual quality upon the second compression.
\item The watermark embedding is implemented in the compressed latent representation of a generic neural compressor. We perform the adversarial perturbations in the compressed domain, which also causes adversarial perturbations in the spatial domain. Any image containing partial unauthorized content is an adversarial example to SLIC.
\item Our SLIC acts as a standard image codec if no watermark is embedded. The watermark embedding is a universal adversarial attack that maps secret messages to perturbations in the compressed domain.
\end{itemize}

\section{Related Works}

\subsection{Adversarial Attacks}

Szegedy et al. \cite{szegedy2013intriguing} first introduced the notion of adversarial examples and demonstrated that neural networks are vulnerable to small perturbations in input data. The goal of an adversarial attack is to find an adversarial example that is indistinguishable from the human eye but can lead to significant misclassifications. Adversarial attacks can be either targeted or untargeted.

Searching for adversarial examples was once considered computationally expensive because it is an optimization problem. Szegedy et al. approximated it using a box-constrained L-BFGS. Goodfellow et al. \cite{goodfellow2014explaining} proposed the Fast Gradient Sign Method (FGSM), a straightforward yet powerful approach for generating adversarial examples. FGSM is computationally efficient because it leverages the gradient of the loss function concerning the input data to create perturbations that maximize the model's prediction error.

Later, more sophisticated attack techniques were introduced. Kurakin et al. \cite{kurakin2018adversarial} extended FGSM by developing the iterative method I-FGSM, which applies FGSM multiple times with small step sizes. Carlini and Wagner \cite{carlini2017towards} proposed the C\&W attacks, which optimized a different objective function to generate adversarial examples. These attacks demonstrated a higher success rate in bypassing defenses.

In response to these attacks, significant efforts \cite{madry2017towards,papernot2016distillation} have been made to propose defense methods. It is known that ``adversarial examples are not bugs but features of neural networks.'' In this work, we propose leveraging adversarial attacks as a strategy to build a secure image codec that destroys itself after re-compression.

\subsection{Learned Image Compression}

The learned image compression technique is an end-to-end approach that automatically learns a pair of image encoders and decoders from a collection of images without the need for hand-crafted design of coding tools. The field of learned image compression has witnessed significant advancements \cite{balle2018variational,minnen2018joint,cheng2020learned,guo2021causal} in the past few years, surpassing traditional expert-designed image codecs. Several comprehensive surveys and introductory papers \cite{ma2019image,yang2022introduction,huang2024unveiling} have summarized these achievements.

However, neural network-based codecs are no exception to adversarial attacks. Kim et al. \cite{kim2020instability} first identified the instability issue of successive deep image compression and proposed including a feature identity loss to mitigate it. Liu et al. \cite{liu2023manipulation} investigated the robustness of learned image compression, where imperceptibly manipulated inputs can significantly increase the compressed bitrate. Such attacks can potentially exhaust the storage or network bandwidth of computing systems. Chen and Ma \cite{chen2023towards} examined the robustness of learned image codecs and investigated various defense strategies against adversarial attacks to improve robustness.

\subsection{DNN-based Data Hiding}

Most DNN-based data hiding methods \cite{zhu2018hidden,zhang2020udh,tancik2020stegastamp,jing2021hinet,luo2020distortion,luo2021dvmark} use their own custom-trained image encoder/decoder for message embedding and hide the invisible perturbations in the spatial domain. Huang and Wu \cite{huang2024joint} first proposed reusing the standard neural codec's encoder/decoder to embed/extract messages in the compressed latent representation. Operating on the compressed latents, their method offers superior image secrecy in steganography and watermarking scenarios compared to existing techniques. Additionally, processing messages in the compressed domain has much lower complexity because additional encoding/decoding is not required.

Once a neural codec's compressed latent representation is jointly trained to allow additional perturbations as hidden data, we can effectively apply perturbations to the latent space as a universal adversarial attack. This compressed domain data hiding method inspires us to develop SLIC as a secure codec.

\subsection{Watermarking to Defend Deepfake}

Adversarial attacks via watermarking are one of the active defense strategies against Deepfakes. Lv \cite{lv2021smart} proposed an adversarial attack-based smart watermark model to exploit Deepfake models. When Deepfake manipulates their watermarked images, the output images become blurry and easily recognized by humans. Yu et al. \cite{yu2021artificial} took a different approach by embedding watermarks as fingerprints in the training data for a generative model. They discovered the transferability of such fingerprints from training data to generative models, thereby enabling Deepfake attribution through model fingerprinting.

Wang et al. \cite{wang2024invisible} proposed an invisible adversarial watermarking framework to enhance the copyright protection efficacy of watermarked images by misleading classifiers and disabling watermark removers. Zhang et al. \cite{zhang2022self} developed a recoverable generative adversarial network to generate self-recoverable adversarial examples for privacy protection in social networks.

\section{Proposed Method}

\begin{figure}[!t]
\centering
\includegraphics[width=1\columnwidth]{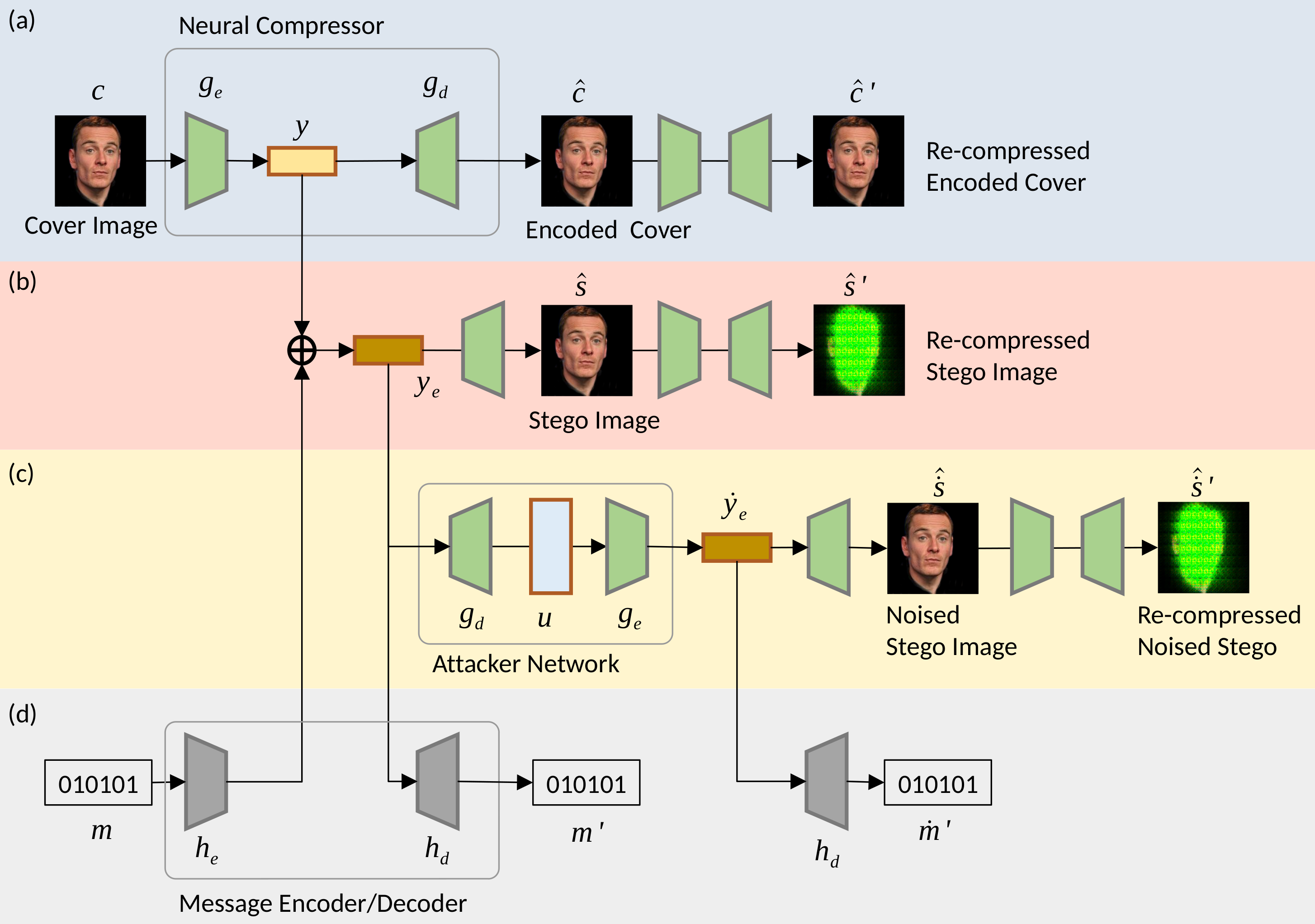}  \\
\caption{The proposed SLIC framework. Our training procedure contains four subflows: (a) image encoding/decoding flow, (b) stego image generating flow, (c) noise attack simulation flow, and (d) message embedding/extraction flow.}
\label{fig:arch}
\end{figure}

We extend the compressed domain data hiding technique \cite{huang2024joint} to learn a secure image codec and reuse its message encoder/decoder architecture. In our SLIC, the latent representation $y$ transformed by the image encoder allows additive small residuals to form the embedded latent code $y_e$. The residuals added to the latent representation are considered a type of adversarial perturbation in the latent space, generating invisible perturbations in the stego image $\hat{s}$ as an adversarial example. If the watermarked stego image is re-compressed, it exploits the image encoder $g_e$ and produces a severely damaged image. We show the proposed SLIC framework in Figure \ref{fig:arch}, which contains four subflows described as follows.

\textbf{Image encoding/decoding flow:} We fine-tune a neural encoder/decoder pair $g_e$ and $g_d$ to stabilize reconstruction quality in the second compression. It is essential that the SLIC must not degrade image quality in the re-compression when an image is not watermarked. Given an input cover image $c$, the first compression by the neural compressor generates an encoded cover image $\hat{c}=g_d(g_e(c))$. Since the cover image is not embedded with a watermark, we expect the re-compressed image $\hat{c}'=g_d(g_e(\hat{c}))$ to be perceptually similar to $\hat{c}$. That is, we aim to minimize:

\begin{equation}
\underset{\theta_e,\theta_d}{\mathrm{arg~min}}~\text{LPIPS}(c,\hat{c}) + \text{LPIPS}(\hat{c},\hat{c}'),
\end{equation}

where $\theta_e$, $\theta_d$ are the parameters of the neural compressor $g_e$,$g_d$ and LPIPS is the DNN-based perceptual loss \cite{zhang2018perceptual}.

\textbf{Stego image generating flow:} We embed a message $m\in \{0,1\}^n$ into a cover image $c$ as a watermark to generate a stego image $\hat{s}$. The compressed latent vector of the cover image is obtained as $y=g_e(c)$, and the embedded latent vector $y_e$ is obtained as follows:

\begin{equation}
y_e=y\oplus h_e(m) \label{eq:embed},
\end{equation}

where $h_e$ is the message encoder that transforms the message to the same dimension as $y$ and $\oplus$ is the element-wise addition. The stego image $\hat{s}=g_d(y_e)$ is obtained by decoding the watermarked latent vector $y_e$ using decoder $g_d$. Here, we optimize the perceptual loss with:

\begin{equation}
\underset{\theta_e,\theta_d}{\mathrm{arg~min}}~\text{LPIPS}(\hat{c},\hat{s}) + \text{ReLU}(\tau - \text{LPIPS}(\hat{s},\hat{s}')),
\end{equation}

where $\hat{s}'=g_d(g_e(\hat{s}))$ is the re-compressed stego image, and $\tau$ is a constant to control the degree of perceptual distance divergence between $\hat{s}$ and $\hat{s}'$. In other words, we expect the stego image to remain visually similar to the encoded cover image without a watermark, while the re-compressed stego image should be perceptually distorted.

\textbf{Noise attack simulation flow:} In practical scenarios, the watermarked stego image may undergo various editing operations such as cropping, rotation, and lighting adjustment. These edits are considered noise attacks on the watermarking system. To simulate noise attacks, we use an attacker $u$ similar to \cite{huang2024joint} and derive the noised stego image as $\dot{s}=u(\hat{s})$, then compress it as $\hat{\dot{s}}=g_d(g_e(\dot{s}))$. Since our SLIC should be robust to noise attacks and damage the visual quality in the re-compression, we optimize the perceptual divergence between $\hat{\dot{s}}$ and $\hat{\dot{s}}'$:

\begin{equation}
\underset{\theta_e,\theta_d}{\mathrm{arg~min}}~\text{ReLU}(\tau - \text{LPIPS}(\hat{\dot{s}},\hat{\dot{s}}')).
\end{equation}

where $\hat{\dot{s}}'=g_d(g_e(\hat{\dot{s}}))$ is the re-compressed noised stego image.

\textbf{Message embedding/extraction flow:} The extracted message $m'=h_d(y_e)$ is obtained with the message decoder $h_d$ on the compressed latent vector $y_e$. The noised embedded latent vector $\dot{y_e}=g_e(\dot{s})$ is then used to extract noised secrets with $\dot{m}'=h_d(\dot{y_e})$. We optimize for the message extraction accuracy by:

\begin{equation}
\underset{\phi_e,\phi_d}{\mathrm{arg~min}}~\text{BCE}(m,m') + \text{BCE}(m,\dot{m}'),
\end{equation}

where $\phi_e$ and $\phi_d$ are the parameters of the message encoder/decoder $h_e$,$h_d$ and BCE is the binary cross-entropy function.

\subsection{Training and Loss Functions} \label{sec:train}

The subflows described in the prior section illustrate the fundamental building blocks of our framework. To learn a secure image codec, we jointly train the image codec and the message encoder/decoder in an interleaved manner, as presented in Algorithm \ref{alg:train}. In each training epoch, we first freeze the image codec parameters and update the message network parameters $\phi_e$ and $\phi_d$ with the data hiding loss function $\mathcal{L}_H$. We then freeze the message network parameters and fine-tune the image codec with the codec loss function $\mathcal{L}_C$.

\begin{algorithm}
    \SetAlgoLined
    \caption{\textbf{SLIC\_Train}($c$)}
    \KwData{Cover image $c$}
    \KwData{Neural encoder/decoder parameters $\theta_e$, $\theta_d$}
    \KwData{Message encoder/decoder parameters $\phi_e$, $\phi_d$ }
    \KwResult{Fine-tuned neural encoder/decoder $g_e$, $g_d$; Trained message encoder/decoder $h_e$, $h_d$}

    \SetKwComment{Comment}{/* }{ */}
    \For{$i \gets 1$ \KwTo $\text{num\_epochs}$}{
        $\mathcal{L}_H \gets \mathcal{L}_P + \alpha \mathcal{L}_M + \beta\mathcal{L}_A$ \;
        $\phi_e \gets \phi_e + lr \times \text{Adam}(\mathcal{L}_H)$ \;
        $\phi_d \gets \phi_d + lr \times \text{Adam}(\mathcal{L}_H)$ \;
        $\mathcal{L}_C \gets \mathsf{R} + \lambda \mathsf{D}$ \;
        $\theta_e \gets \theta_e + lr \times \text{Adam}(\mathcal{L}_C)$ \;
        $\theta_d \gets \theta_d + lr \times \text{Adam}(\mathcal{L}_C)$ \;
    }
    \Return\;
\label{alg:train}
\end{algorithm}

All the optimization objectives in each subflow are reformulated using different loss functions during training. We design the data hiding loss function as a combination of perceptual loss $\mathcal{L}_P$, message loss $\mathcal{L}_M$, and adversarial loss $\mathcal{L}_A$:

\begin{equation}
\mathcal{L}_H = \mathcal{L}_P + \alpha \mathcal{L}_M + \beta\mathcal{L}_A,
\end{equation}

where $\alpha$ and $\beta$ are hyper-parameters used to control the relative weights. We measure the perceptual loss with the DNN-based perceptual metric LPIPS \cite{zhang2018perceptual}:

\begin{equation}
\mathcal{L}_P = \text{LPIPS}(\hat{c},\hat{s}) + \text{LPIPS}(\hat{c},\hat{c}').
\end{equation}

The second term of $\mathcal{L}_P$ is designed to ensure an image with no watermark will remain perceptually stable after re-compression. We avoid using MSE (mean square error) to minimize image distortion because we observed that the LPIPS metric significantly impacts our standard neural codec more than a custom-trained image encoder/decoder. The ablation study is provided in Appendix~A.

To measure the decoded message error, we define the message loss function as:

\begin{equation}
\mathcal{L}_M = \text{BCE}(m,m') + \gamma \text{BCE}(m,\dot{m}'),
\end{equation}

where $\gamma$ is a hyper-parameter that controls the weight of the noised message error term. The adversarial loss is designed to make the stego image $\hat{s}$ and its noised version $\hat{\dot{s}}$ become adversarial examples to the neural image codec in the re-compression:

\begin{equation}
\mathcal{L}_A = \text{ReLU}(\tau - \text{LPIPS}(\hat{s},\hat{s}')) + \text{ReLU}(\tau - \text{LPIPS}(\hat{\dot{s}},\hat{\dot{s}}')),
\end{equation}

where $\tau$ is a constant threshold to enforce perceptual distance divergence between the original and re-compressed stego images.

The codec loss $\mathcal{L}_C$ optimizes the rate-distortion efficiency by minimizing:

\begin{equation}
\mathcal{L}_C = \mathsf{R} + \lambda \mathsf{D},
\end{equation}

where $\lambda$ is a hyper-parameter that controls the trade-off between rate and distortion.

For any image $x \in \mathcal{X}$, the neural encoder $g_e$ transforms $x$ into a latent representation $y=g_e(x)$, which is later quantized to a discrete-valued vector $\hat{y}$. To reduce the notational burden, we refer to $\hat{y}$ simply as $y$ in what follows. The discrete probability distribution $P_y$ is estimated using a neural network and then encoded into a bitstream using an entropy coder. The \textit{rate} of this discrete code, $\mathsf{R}$, is lower-bounded by the entropy of the discrete probability distribution $H(P_y)$. Because the perturbations (i.e., hidden message) added to the latent code break the original entropy coding optimality, we revise the rate function to include both $y$ and $y_e$ when we freeze the message network and fine-tune the image codec. That is:

\begin{equation}
\mathsf{R} = H(P_{y}) + H(P_{y_e}).
\end{equation}

On the decoder side, we decode $y$ from the bitstream and reconstruct the image $\hat{x}=g_d(y)$ using the neural decoder. The \textit{distortion}, $\mathsf{D}$, is measured by a distance metric $d(x,\hat{x})$, where mean square error is commonly used.

\subsection{Noise Attacks}

We define eight types of editing operations commonly used in creating spliced fake images, including copy, Gaussian blur, median filtering, lightening, sharpening, histogram equalization, affine transform, and JPEG compression. We then test the resilience of our stego image against these image manipulations and evaluate the effectiveness of SLIC upon re-compression.

Surprisingly, our SLIC demonstrates robustness against non-filtering editing operations such as lightening, sharpening, and histogram equalization, even without noise attack simulation. These attacks tend to increase rather than decrease the magnitude of the adversarial perturbation. We classify Gaussian blur, median filtering, affine transform, and JPEG compression as filtering editing operations because the adversarial perturbations added to the stego image through latent space perturbations are partially filtered due to resampling. Therefore, to enhance robustness, we simulate two types of noise attacks during training:

\textbf{Affine transform:} We randomly rotate from -10 to 10 degrees, translate from 0\% to 10\% in both axes and scale from 90\% to 110\% on the stego samples.

\textbf{JPEG compression:} We re-compress the stego image with random JPEG quality from 70 to 95.

We do not simulate the blurring attack because the affine transform has a similar low-pass filtering effect from rotation and scaling. Further discussion on this is provided in Section \ref{sec:adv_effect}.

\section{Experimental Results} \label{sec:experiment}

\begin{table*}[!t]
\centering
\caption{The quality metrics comparison of the SLIC with different neural codecs on test datasets.}
\resizebox{\textwidth}{!}{
\begin{tabular}{lrr|rrrrrrrrrr}
\hline
\multirow{2}{*}{\textbf{SLIC}} & \multirow{2}{*}{\textbf{PSNR}$(\hat{s})\uparrow$} & \multirow{2}{*}{\textbf{BER}$\downarrow$} & \multicolumn{10}{c}{\textbf{Re-compressed PSNR}} \\ 
 & & & \textbf{Cover$(\hat{c}')\uparrow$} & \textbf{Stego$(\hat{s}')\downarrow$} & \textbf{Copy$\downarrow$} & \textbf{G.Blur$\downarrow$} & \textbf{M.Filter$\downarrow$} &  \textbf{Sharp$\downarrow$} & \textbf{Light$\downarrow$} & \textbf{H.Equ.$\downarrow$} & \textbf{Affine$\downarrow$} & \textbf{JPEG$\downarrow$} \\ \hline
\multicolumn{13}{c}{\textbf{Kodak}} \\
\hline
Balle2018 & 42.58 & 0.012 & 40.75 & 6.38 & 9.60 & 7.33 & 7.60 & 5.60 & 6.94 & 5.60 & 28.17 & 26.05 \\
Minnen2018 & 41.08 & 0.005 & 42.14 & 6.09 & 9.41 & 8.34 & 8.21 & 4.90 & 6.62 & 4.82 & 21.22 & 28.81 \\
Cheng2020 & 41.03 & 0.008 & 43.23 & 5.78 & 6.63 & 34.22 & 38.14 & 4.90 & 7.63 & 4.94 & 20.86 & 6.35 \\
\hline
\multicolumn{13}{c}{\textbf{DIV2K}} \\
\hline
Balle2018 & 41.30 & 0.023 & 40.67 & 6.52 & 9.91 & 8.06 & 8.48 & 6.08 & 8.23 & 5.90 & 31.67 & 22.38 \\
Minnen2018 & 39.60 & 0.010 & 43.41 & 5.87 & 8.84 & 9.08 & 9.06 & 4.49 & 7.48 & 5.16 & 23.93 & 25.74 \\
Cheng2020 & 40.08 & 0.041 & 40.91 & 5.11 & 6.44 & 32.39 & 36.74 & 4.31 & 6.74 & 4.94 & 22.40 & 5.86 \\
\hline
\multicolumn{13}{c}{\textbf{CelebA}} \\
\hline
Balle2018 & 43.66 & 0.006 & 39.06 & 5.61 & 8.91 & 6.70 & 6.73 & 4.73 & 8.31 & 5.41 & 29.81 & 26.47 \\
Minnen2018 & 41.98 & 0.001 & 45.00 & 5.65 & 8.42 & 7.98 & 7.41 & 4.55 & 7.25 & 4.92 & 21.62 & 32.13 \\
Cheng2020 & 41.43 & 0.003 & 44.50 & 5.27 & 6.30 & 33.77 & 36.02 & 4.68 & 6.74 & 4.92 & 21.63 & 5.61 \\
\hline
\end{tabular} \label{tab:stego-distortion}
}
\end{table*}

We implemented our SLIC using the CompressAI \cite{begaint2020compressai} release of neural codecs \cite{balle2018variational,minnen2018joint,cheng2020learned}, denoted as Balle2018, Minnen2018, and Cheng2020. For training, we randomly selected 90\% of the images from the COCO dataset \cite{russakovsky2015imagenet} as the training set, and the remaining 10\% as the validation set. Cover images were resized to $128 \times 128$, and 64-bit binary messages were randomly embedded during training. We evaluated our SLIC on the Kodak \cite{kodakcd}, DIV2K \cite{Agustsson_2017_CVPR_Workshops}, and CelebA \cite{liu2015deep} datasets.

During training, we employed the PyTorch built-in Adam optimizer with a learning rate of 0.001 for the message network and a batch size of 32. Each neural image codec was fine-tuned with a learning rate of $5 \times 10^{-5}$. The training was performed for 160 epochs. Hyperparameters were set as follows: $\tau=2.0$, $\alpha=1.5$, $\beta=1.0$, and $\gamma=1.0$. Our experiments were conducted on an Intel i7-9700K workstation with an Nvidia GTX 3090 GPU.

For the test editing operations, we configured them as follows: copying the stego image onto another larger blank image, Gaussian blur with a $3 \times 3$ window and $\sigma=1.0$, median filtering with a $3 \times 3$ window, sharpening with a $3 \times 3$ kernel, a 50\% increase in luminance in the LAB color space, and histogram equalization in the RGB channel. We rotated by 5 degrees, translated by 5 pixels, and scaled to 95\% for affine transform. For JPEG compression, we compressed the stego image with a quality of 80.

\subsection{Watermarked Image Quality} \label{sec:secrecy}

Quantitatively, we present the SLIC's watermarked image quality in PSNR metric and the bit error rate (BER) in Table \ref{tab:stego-distortion}. The PSNR of watermarked images ($\hat{s}$) compared to images without watermark ($\hat{c}$) is around 40 for all three neural codecs, which remains suitable for visual communication. The bit error rate of SLIC is unable to achieve zero, but there are established error correction techniques \cite{bose1960class,choi2019neural} to mitigate.

We report the re-compression effects in PSNR in Table \ref{tab:stego-distortion}, including the cover, stego, and stego images with eight pre-defined editing operations. In the third column of Table \ref{tab:stego-distortion}, the PSNR of the re-compressed cover images stays close to that of the watermarked image, as expected because the re-compression should not cause any quality degradation on images without watermarks. The fourth column shows that the watermarked image's quality drops significantly after re-compression, where the PSNR is around 6 on average. Even with image editing such as Gaussian blur and median filtering, the watermarked SLIC images still cause severe quality degradation in a second compression.

Since adversarial stego images are the result of invisible perturbations added to the pixels, the affine transform and JPEG compression could effectively reduce those perturbations. Table \ref{tab:stego-distortion} shows that the re-compressed stego images after affine and JPEG attacks have a PSNR of around 25.

According to \cite{huang2024joint}, a codec with lower coding efficiency, such as Balle2018, tends to have higher stego image quality because its latent space has more room for data hiding. However, the relation between coding efficiency and re-compression quality degradation needs to be clarified, as indicated in Table \ref{tab:stego-distortion}. Another interesting finding is that Cheng2020 seems to have different patterns of adversarial effects in Gaussian blur, median filter, and JPEG compression. We think the cause may be the attention module in Cheng2020, which requires further investigation.

\subsection{Adversarial Effects} \label{sec:adv_effect}

Qualitatively, we present the re-compressed results of altered stego images in Figure \ref{fig:re-compress3}. More adversarial effects regarding codecs Minnen2018 and Cheng2020 are provided in Appendix~B.

\begin{figure}[!ht]
\centering
\includegraphics[width=\columnwidth]{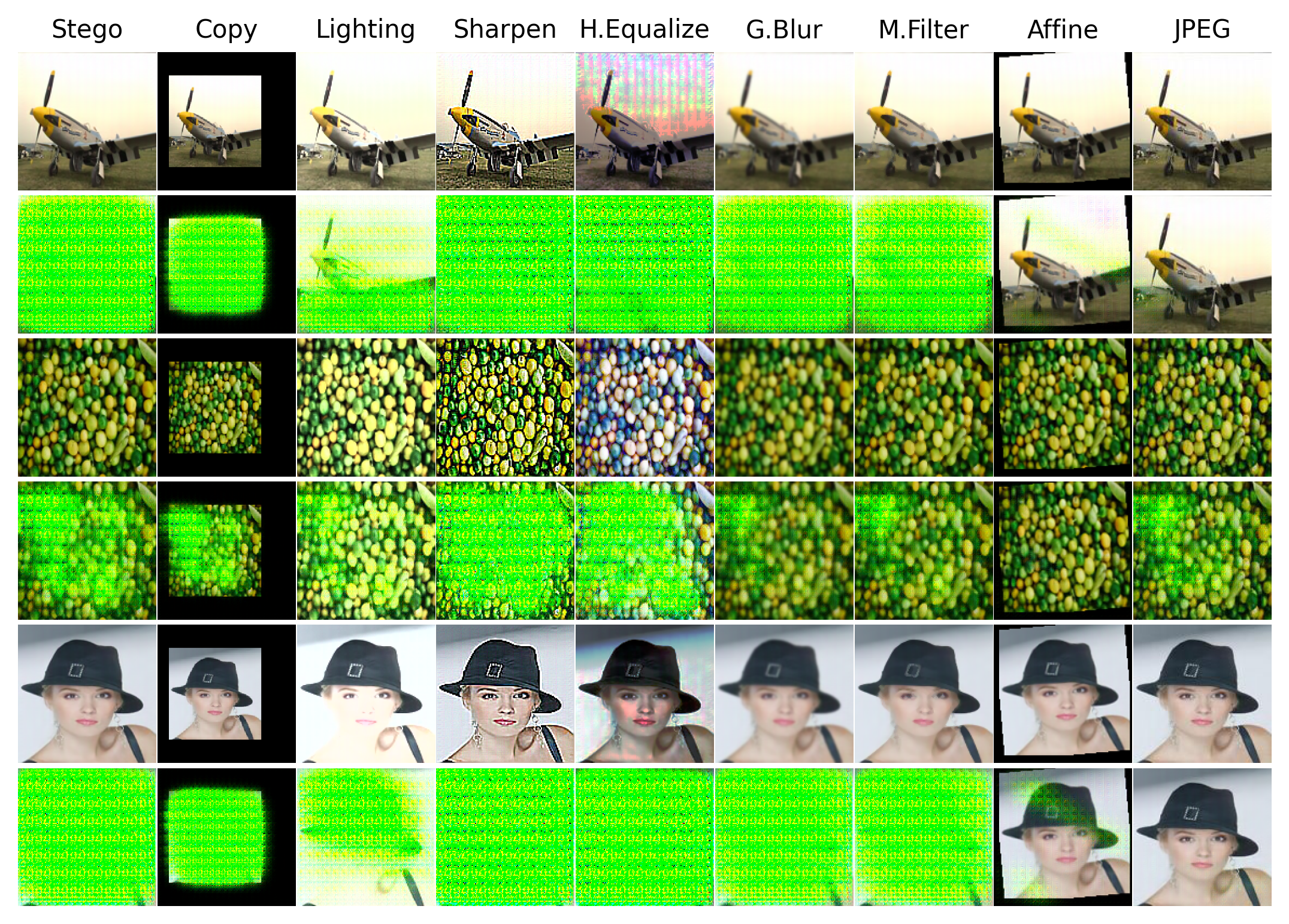}
\caption{The re-compressed results of stego images with various editing operations. The SLIC is based on the Balle2018 codec.}
\label{fig:re-compress3}
\end{figure}

In Figure \ref{fig:re-compress3}, the first and second columns are the original stego image and the stego image copied onto another blank image, respectively. Both of them were seriously damaged in the second compression, as we expected. Because our SLIC achieves the secure objective via adding adversarial perturbation in the latent space of the stego image, image manipulations that magnify the perturbation, such as lightening, sharpening, and histogram equalization, will not break the adversarial attack.

Following that in Figure \ref{fig:re-compress3} are filtering operation results. Although the adversarial perturbations are partially filtered due to resampling, our SLIC with the noise attack simulation can still severely degrade the image quality after Gaussian blur and median filtering editing. Though the affine and JPEG-attacked images are still visually applicable to humans, their quality drops are easy to spot due to some generated artifacts shown in Figure \ref{fig:re-compress3}.

\subsection{Watermark Robustness}

\begin{figure}[!h]
\centering
\includegraphics[width=1\columnwidth]{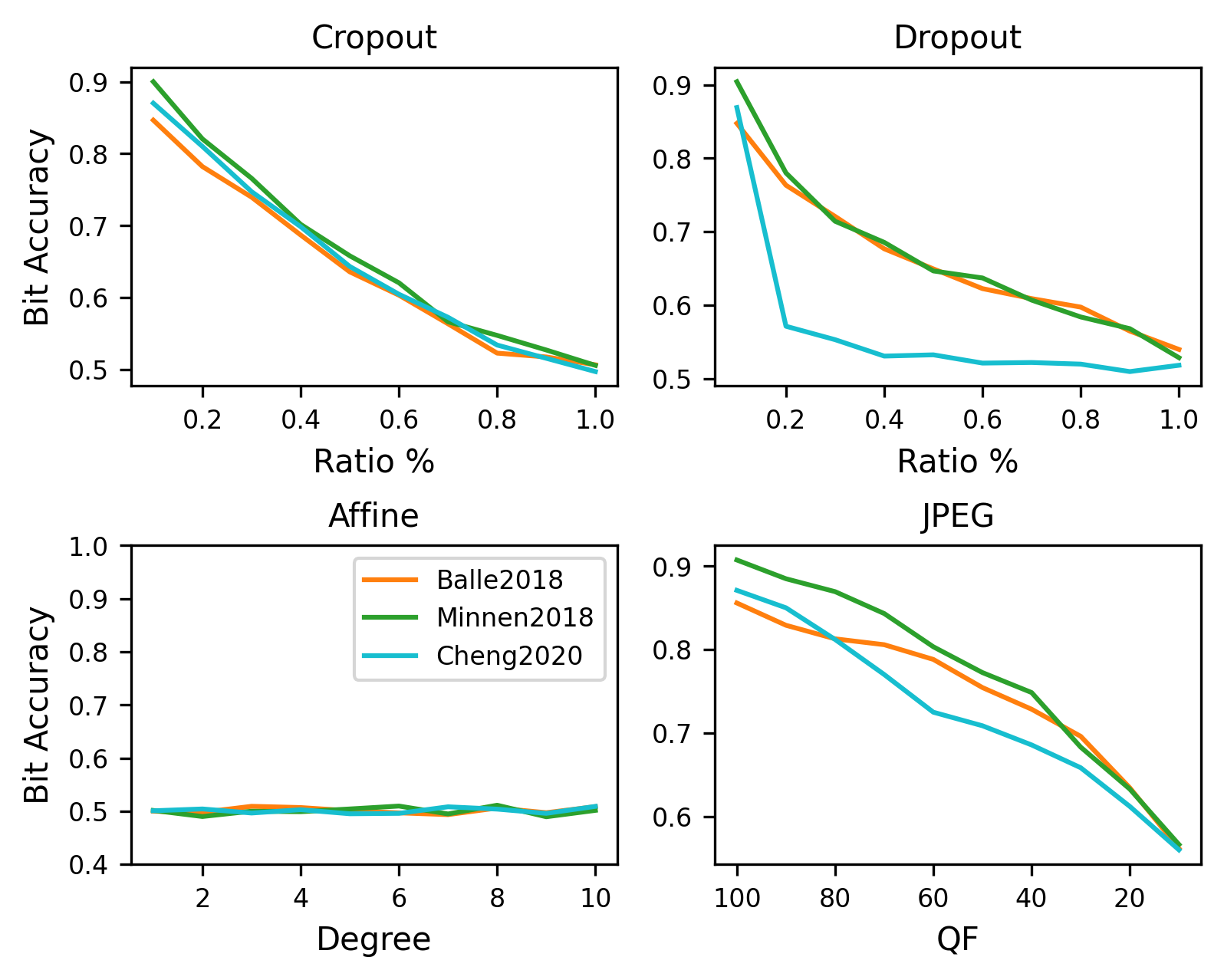}
\caption{Watermark robustness against selected noise attacks evaluated on DIV2K.}
\label{fig:noise-robust}
\end{figure}

We evaluated the watermark robustness of our SLIC against noise attacks cropout, dropout, affine, and JPEG on the DIV2K dataset, as shown in Fig. \ref{fig:noise-robust}. We varied the attack strength by increasing the noise parameter, which degrades image quality along the horizontal axis. We do not add the cropout and dropout attack simulations during training; we merely use affine and JPEG as proxies. From the watermark extraction point of view, the SLIC maintains robustness against selected attacks except for affine transform, which is generally considered challenging in the watermarking field. However, the content owner can re-compress the protected image with SLIC to spot the quality degradation as another piece of evidence of infringement.

\subsection{Coding Efficiency Impact}

Conceptually, watermarking techniques convert secret messages to noise-like signals and add them onto cover images as invisible perturbations. The embedding operation will increase the entropy of the stego image and the file size after image compression. Table \ref{tab:overhead} shows the stego image file size overhead by percentage. As mentioned in Section \ref{sec:train}, our SLIC jointly optimizes the data hiding process and coding efficiency. Thus, the produced stego image size has a negligible overhead, around 1-2\%. Our fine-tuned neural codec as SLIC remains optimal in coding efficiency compared to vanilla ones.

\begin{table}[!ht]
\caption{Watermark Embedding overhead}
\label{tab:overhead}
\centering
\begin{tabular}{lrrr}
\hline
\textbf{SLIC} & \textbf{Kodak} & \textbf{DIV2K} & \textbf{CelebA} \\ \hline
Balle2018 & 1.37\% & 1.15\% & 2.37\% \\
Minnen2018 & 1.16\% & 0.77\% & 1.39\% \\
Cheng2020 & 0.88\% & 0.87\% & 1.71\% \\
\hline
\end{tabular}
\end{table}

\section{Conclusion}

This paper presents the Secure Learned Image Codec (SLIC), a novel active approach to ensuring image authenticity. SLIC effectively combats unauthorized image manipulation by embedding watermarks in the compressed domain, generating adversarial examples that degrade visually upon re-compression. Our method introduces an adversarial loss to train the SLIC, ensuring that watermarked images suffer noticeable quality degradation after re-compression if tampered with.

Experimental results demonstrate SLIC's robustness against various noise attacks and common image editing operations, with significant artifacts appearing in counterfeit images, thus preventing their redistribution. This work represents an initial attempt to develop a secure image codec with the potential for widespread adoption once standardized and trusted by the public. Future research should explore the adversarial effects of different neural codecs and enhance noise attack simulations to further deteriorate image quality after editing and re-compression. By refining these techniques, we can advance towards more resilient defenses against digital image counterfeiting, ultimately protecting the integrity of digital media from the outset.

\begin{acks}
The authors would like to thank the NSTC of Taiwan and National Taiwan University for supporting this research under the grant numbers 111-2221-E-002-134-MY3 and NTU-112L900902.
\end{acks}

\bibliographystyle{ACM-Reference-Format}
\bibliography{all_refs}

\newpage
%
%
\appendix

\section{Perceptual Loss Study} \label{sec:ablation}

The success of message extraction is highly dependent on the design of the message encoder/decoder when jointly trained with a standard codec that minimizes image distortion. Experimental results show that the widely used MSE loss performs poorly. The recovered bit accuracy is good, but the stego image has been visually deteriorated, as shown in preset $\mathrm{a}$ of Table \ref{tab:ablation-lpips}.

To address this issue, we added LPIPS in the perceptual loss function in preset $\mathrm{b}$, demonstrating a significant improvement in image quality. Compared to a custom-trained image encoder/decoder, the LPIPS metric significantly impacts the reconstruction quality of our standard neural codec. This observation could be due to the high compactness of our neural codec, which is more strongly connected to certain perceptual features in the latent space. Consequently, adding LPIPS as a distortion loss function can effectively optimize the message encoder to modify the compressed latents. Finally, we propose using the preset $\mathrm{c}$, which omits the MSE loss, as our proposed perceptual loss function.

\begin{table}[!ht]
\caption{Distortion comparison with different $\mathcal{L}_P$ loss}
\label{tab:ablation-lpips}
\centering
\begin{tabular}{lrrrrr}
\hline
\multicolumn{5}{c}{\textbf{DIV2K}} \\ \hline
\textbf{SLIC} & \textbf{PSNR} & \textbf{SSIM} & \textbf{MAE} & \textbf{$\text{LPIPS}(\hat{c},\hat{s})$}  \\ \hline
Balle2018$^{\mathrm{a}}$ & 31.65 & 0.9333 & 4.87  & 0.01638 \\
Minnen2018$^{\mathrm{a}}$ & 31.07 & 0.9288 & 5.31 & 0.02358 \\ \hline
Balle2018$^{\mathrm{b}}$ & 41.55 & 0.9943 & 1.41 & 0.00065 \\
Minnen2018$^{\mathrm{b}}$ & 38.31 & 0.9840 & 2.32 & 0.00465 \\ \hline
Balle2018$^{\mathrm{c}}$ & 41.67 & 0.9945 & 1.36 & 0.00064 \\
Minnen2018$^{\mathrm{c}}$ & 38.42 & 0.9847 & 2.25 & 0.00485 \\ \hline
\multicolumn{5}{l}{$^{\mathrm{a}}\mathcal{L}_P=\text{MSE}(\hat{c},\hat{s})$} \\
\multicolumn{5}{l}{$^{\mathrm{b}}\mathcal{L}_P=\text{MSE}(\hat{c},\hat{s}) + \text{LPIPS}(\hat{c},\hat{s})$} \\
\multicolumn{5}{l}{$^{\mathrm{c}}\mathcal{L}_P=\text{LPIPS}(\hat{c},\hat{s}) $} \\
\end{tabular}
\end{table}

\section{More Adversarial Effects} \label{sec:adv_imgs}

\begin{figure*}[!ht]
\centering
\includegraphics[height=0.42\textheight]{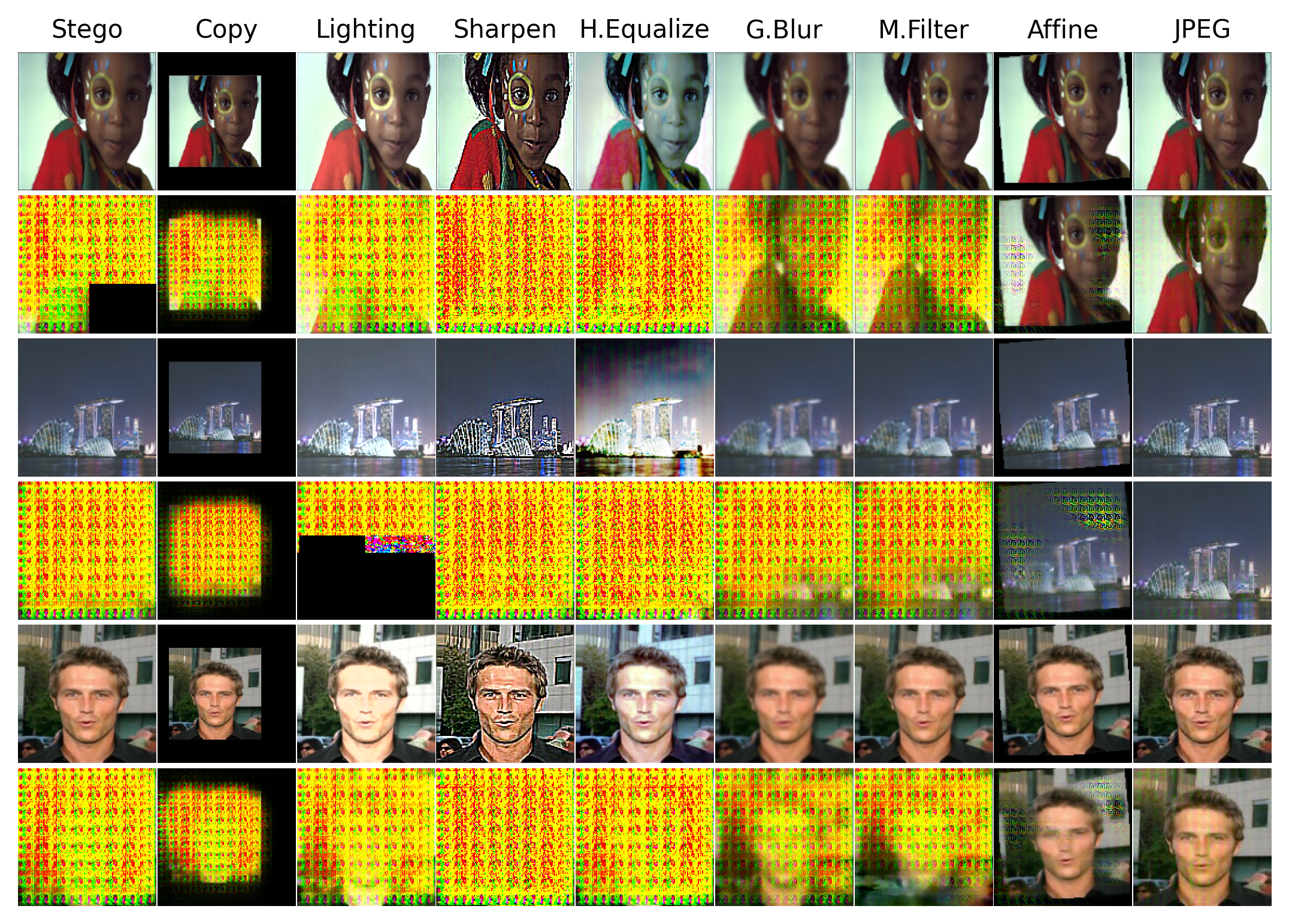}
\caption{The re-compressed results of stego images with various editing operations. The SLIC is based on the Minnen2018 codec.}
\label{fig:re-compress4}
\end{figure*}

\begin{figure*}[!ht]
\centering
\includegraphics[height=0.42\textheight]{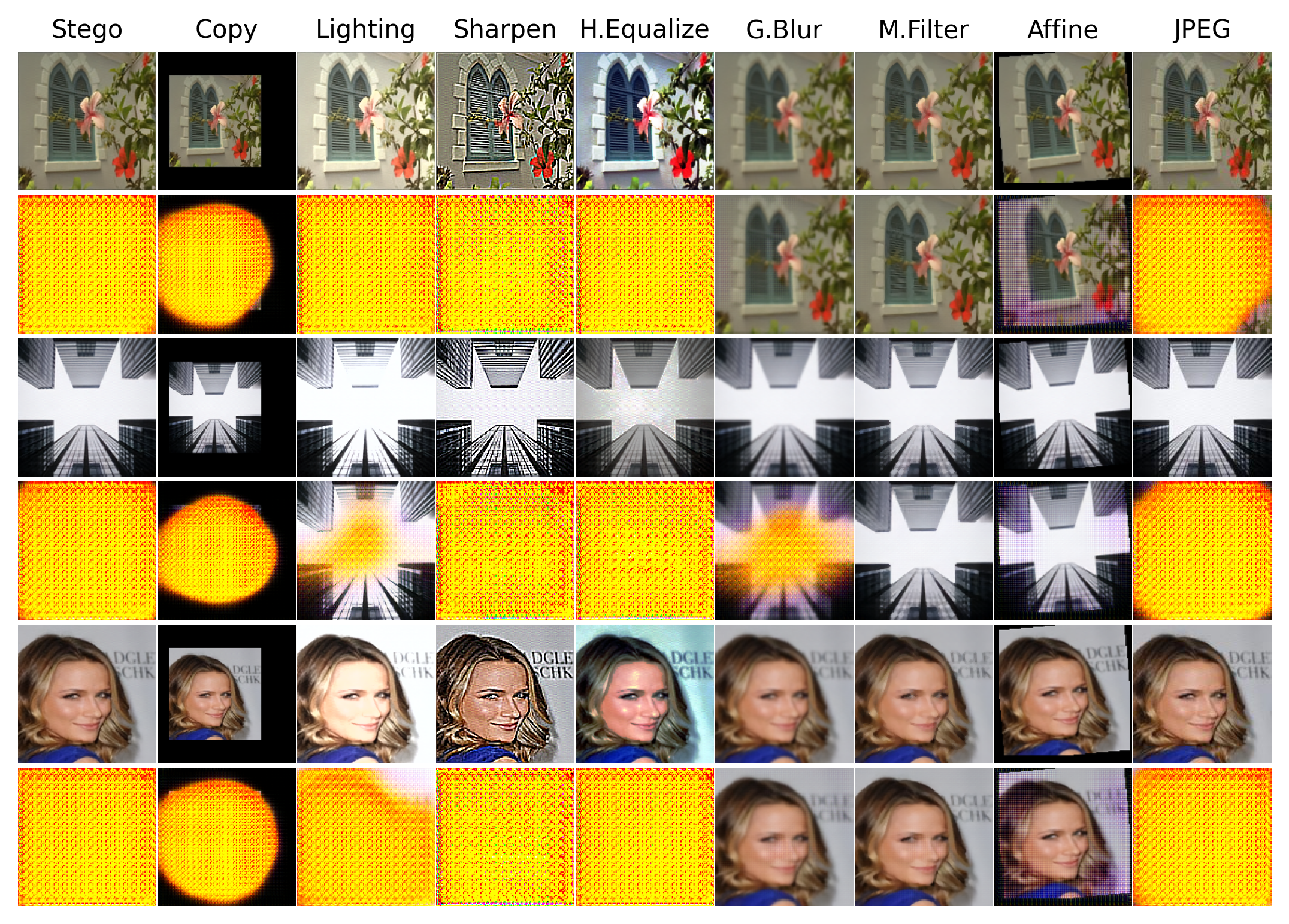}
\caption{The re-compressed results of stego images with various editing operations. The SLIC is based on the Cheng2020 codec.}
\label{fig:re-compress5}
\end{figure*}

We present more qualitative results of adversarial stego images in Figures \ref{fig:re-compress4} and \ref{fig:re-compress5} based on neural image codec Minnen2018 and Cheng2020, respectively.

\end{document}